\documentclass[11pt]{article}

\usepackage[]{acl}
\usepackage{tcolorbox}

\usepackage{times}
\usepackage{latexsym}
\usepackage{enumitem}
\usepackage[table,xcdraw]{xcolor}
\usepackage{makecell}
\usepackage[T1]{fontenc}

\usepackage[utf8]{inputenc}

\usepackage{microtype}

\usepackage{inconsolata}
\usepackage{multirow} 

\usepackage{graphicx}

\title{Can LVLMs Uncover the Truth Behind Visual Illusions? An Analysis of Perceptual and Reasoning Capabilities
}

\author{Liangjie Zhao\textsuperscript{\rm1,2}\space\space 
Jiaqing Lyu\textsuperscript{\rm 3}\space\space 
\textbf{Kexin Tang}\textsuperscript{\rm 4}\thanks{Project Leader.}\space\space 
\textbf{Zecheng Fang}\textsuperscript{\rm 2}  \space\space \\
\textbf{Rong Yin}\textsuperscript{\rm 5}\space\space 
\textbf{Yulan Hu}\textsuperscript{\rm 6}\space\space 
\textbf{Da Li}\textsuperscript{\rm 1,2}\thanks{Corresponding authors.}\space\space
\textbf{Jianing Li}\textsuperscript{\rm 1}\footnotemark[\value{footnote}]\space\space \\
\textsuperscript{\rm 1}State Key Laboratory of AI Safety, Institute of Computing Technology, \\
Chinese Academy of Sciences \,
\textsuperscript{\rm 2}University of Chinese Academy of Sciences\\
\textsuperscript{\rm 3}Tsinghua University,\,
\textsuperscript{\rm 4}Amap, Alibaba Group,\,
\textsuperscript{\rm 5}Beihang University,\,
\textsuperscript{\rm 6}Renmin University\\
zhaoliangjie55@gmail.com,\,\, jixuan.tkx@alibaba@inc.com,\,\,yinrong@buaa.edu.cn\,\,\\
huyulan@ruc.edu.cn,\,\,lida21s@ict.ac.cn,\,\,lijianing@ict.ac.cn\\
}

\newcommand{\benchmarkname}{IllusionReasoning}

\begin{document}
\maketitle
\begin{abstract}
Large Vision Language Models have integrated reasoning capabilities, elevating cognitive performance to new levels. However, existing evaluations either focus solely on perception or rely on specific domains such as maths or coding. Evaluation for reasoning capabilities that align with an open-world environment is still required, especially one that considers perception and reasoning jointly. To bridge this gap, we propose to evaluate LVLMs by exploiting visual illusions as a diagnostic tool. Visual illusions are phenomena in which the human visual system misinterprets objective signals, resulting in an understanding that deviates from reality. We constructed \textbf{\benchmarkname},  a benchmark of illusion images collected from the real world, incorporating diverse annotated question-answer pairs. Based on \benchmarkname, we show that the reasoning capabilities of a wide range of LVLMs are not as advanced as claimed. Our work provides new insights into LVLMs and offers future direction for optimisation. Our project is publicly available at
\url{https://github.com/zhaoliangjie55/EMNLP2026_Illusion}.
\end{abstract}

\section{Introduction}
Large Vision Language Models (LVLMs)~\citep{openai2024gpt4o,bai2025qwen3vltechnicalreport} demonstrate outstanding performance in tasks such as visual question answering and multimodal dialogue by integrating visual perception modules with the language modeling capabilities of Large Language Models~\citep{meta2025llama4,openai2024gpt4technicalreport,yang2025qwen3technicalreport}. 
Building upon this foundation, LVLMs further incorporate reasoning through strategies such as Chain-of-Thought and Reinforcement Learning, achieving human-like thinking capabilities exemplified by models such as OpenAI-o1~\citep{openai2024openaio1card} and DeepSeek-R1~\citep{deepseekai2025deepseekr1incentivizingreasoningcapability}.
This integration enables LVLMs to tackle complex tasks such as logical visual question answering, pushing their capabilities to new levels.

With the improvement of capabilities, the evaluation of LVLMs is shifting toward pursuing comprehensive competitiveness in multi-tasking~\citep{liu2024mmbenchmultimodalmodelallaround,mmstar,simplevqa}.
However, these evaluation frameworks originally focus on perception. As LVLMs such as Qwen3-VL~\citep{bai2025qwen3vltechnicalreport} integrate reasoning capabilities, the evaluation starts to focus on their reasoning performance on complex tasks such as OlympiadBench~\citep{OlympiadBench} and MathVista~\citep{mathvista}. However, such domain-specific evaluations are insufficient to fully capture the genuine reasoning capabilities of LVLMs. The core reason lies in their limited coverage beyond these specific domains. The requirement for paradigms that closely align with the complex logic scenarios in open-world environments is urgent, in order to better evaluate the reasoning capabilities of LVLMs. Visual illusions naturally satisfy these requirements.

Visual illusions~\citep{gregory1968perceptual, gregory1997knowledge, bach2006optical} refer to phenomena where visual perception conflicts with cognitive reality, fundamentally stemming from biases within the human intuition.
As a prevalent scenario of visual cognitive conflict in open worlds, visual illusions encompass diverse inputs from real-world settings.
Addressing illusion-related questions requires LVLMs to invoke cross-modal common sense, causal logic, and cognitive reasoning capabilities to discern the underlying logic behind visual phenomena~\citep{sun2021imagenet}. There has been some work focusing on illusion-related content as an evaluation benchmark for LVLMs. These studies typically select classic visual illusions such as the Müller-Lyer line illusion~\citep{garcia2015muller} and the Penrose triangle spatial paradox~\citep{plotnitsky1997penrose}. By designing questions such as ``Are the two line segments in the image actually equal in length?'' or ``Could this three-dimensional figure exist in real physical space?'', these studies use the output to evaluate whether LVLMs can overcome misleading visual appearances rather than rely solely on raw visual features to conclude~\citep{zhang2025illusionbenchlargescalecomprehensivebenchmark,illusionvqa,zhang2023grounding}.

The aforementioned evaluations focus primarily on the perceptual capabilities of LVLMs. In this work, we attempt to directly analyze the perceptual and reasoning capabilities of LVLMs through visual illusions. First, we constructed \textbf{\benchmarkname}, a benchmark of visual illusions from the real world. Then, based on these images, we encourage human annotators to raise questions and provide answers across the following dimensions: (1) \textit{Detection}: Given an object, can LVLMs determine whether it appears in an image? (2) \textit{Description}: How do LVLMs describe illusion images? (3) \textit{Causes of Illusions}: When informed about the content of visual illusions, can LVLMs explain the causes? After \benchmarkname~is built, we aim to evaluate LVLMs from different perspectives: (1) \textbf{Visual Object Recognition}: Do LVLMs possess basic image understanding capabilities? (2) \textbf{Alignment Preference}: Do LVLMs prefer human intuition or the physical world? 
(3) \textbf{Reasoning}: Can LVLMs analyze the causes of inconsistency between the illusion and the physical world?
Among these perspectives, the first two are designed to evaluate the perceptual capabilities, while the last one is intended to evaluate reasoning capabilities.

We evaluated a wide range of LVLMs on \benchmarkname, encompassing open-source and closed-source models of varying sizes. We found that even state-of-the-art LVLMs struggle to provide correct responses to questions related to visual illusions. 
We also conducted a comprehensive analysis from multiple perspectives, including question categories, alignment preferences, and reasoning capabilities. 
We found that LVLMs do not consistently benefit from their thinking mode across different tasks. In simple reasoning-related tasks, excessive thinking may harm performance. Furthermore, when faced with hard problems, LVLMs tend to generate safe responses to avoid making claims on ambiguous content. This harms performance for illusion identification and may hardly be avoided under existing alignment paradigms. Refined strategies are needed to mitigate this phenomenon and enhance the capabilities of LVLMs.

In summary, our contributions are as follows:
\begin{itemize}[leftmargin=*]
\item We collected and constructed \benchmarkname~, a benchmark of visual illusions from the real world to evaluate the perception and reasoning capabilities of LVLMs.
\item We conducted a comprehensive evaluation of widely used LVLMs. The results indicate that the reasoning capabilities of these models are not as good as claimed.
\item We analyzed the perception and reasoning capabilities of LVLMs from multiple dimensions, providing direction for subsequent optimisation.
\end{itemize}

\section{Related Work}
\subsection{Large Visual-Language Model}
By combining the semantic reasoning capabilities of Large Language Models (LLMs) with the perceptual capabilities of visual encoders, Large Visual-Language Models (LVLMs) deliver strong performance in visual question answering and embodied intelligence~\citep{lu2024deepseek,huang2024llm2clip}. LVLMs typically consist of three components: a visual encoder, a connector, and an LLM.
As the core of LVLMs, visual encoders, typically based on ViT~\citep{dosovitskiy2020image}, process raw visual inputs into high-dimensional visual features. Then, a cross-modal alignment connector is required to project the visual features from the visual encoder into the space of the text and enable the LLM to interpret visual data as textual tokens~\citep{liu2023visual, chen2024internvl, zhao2025beyond,zhu2023minigpt}. This transformation empowers LLMs with visual capabilities, enabling them to complete vision-related tasks.

\subsection{Evaluation of Hallucination}
Hallucinations pose a major obstacle to the application of LVLMs in real-world scenarios. Existing benchmarks have evolved from simple object-level evaluations to more complex evaluations of attributes and relations~\cite{bai2024hallucination}. POPE~\citep{li2023evaluating} established a benchmark for object existence, while AMBER~\cite{wang2023amber} extends the evaluation scope to measure hallucinations regarding attributes and relationships. Recently, PhD~\citep{liu2025phdchatgptpromptedvisualhallucination} introduced a ChatGPT-prompted benchmark design and expanded the evaluation perspectives. However, these hallucination benchmarks primarily focus on perceptual capabilities and lack a design tailored to the evaluation of reasoning capabilities.

\subsection{Evaluation of Illusion}

Illusion-induced errors in LVLMs are considered as a special type of hallucination, and few works have specifically evaluated illusion. GVIL~\cite{zhang2023grounding} pioneered this by assessing human alignment across five classic illusion types. HallusionBench~\cite{guan2024hallusionbench} introduced visual control groups to decouple visual misperception from language priors. IllusionVQA~\cite{illusionvqa} expanded the scope to 12 categories, incorporating soft localization for geometric inconsistencies. More recently, IllusionBench+~\cite{zhang2025illusionbenchlargescalecomprehensivebenchmark} scaled up evaluation with ``Trap Illusions'' and colorblindness tests to identify shortcut learning behaviors. Despite these advancements, existing datasets primarily rely on fixed question formats, allowing models to exploit random guessing or language shortcuts without demonstrating true visual comprehension. Furthermore, regarding the visual data itself, recent mechanism studies~\cite{ullman2024illusionillusionvisionlanguagemodels, shinozaki2025largevisionlanguagemodelsdistinguish} use ``anti-illusions'' to confirm that illusion errors are primarily prior-driven reasoning errors rather than perception errors. This implies that models heavily rely on memorized knowledge when processing classic or synthetic illusions. Consequently, evaluating on these synthetic images fails to accurately measure genuine visual perception and reasoning.
\section{Why Do We Need \benchmarkname?}
\subsection{Model Capabilities: From Perception to Reasoning}
The evolution of LVLMs demonstrates a transition from perceptual processing to sophisticated reasoning capabilities. LVLMs, exemplified by LLaVA~\citep{liu2023visual} and InternVL~\citep{chen2024internvl}, have acquired foundational perception capabilities and can tackle tasks such as image-text matching and simple visual question-answering. 
However, they remain confined to a superficial understanding of modal information. With breakthroughs in text reasoning by LLMs, some studies began exploring the adaptation of the Chain-of-Thought (CoT) reasoning paradigm to visual-language tasks. This adaptation gave rise to models such as LLaVA-CoT~\citep{xu2025llavacotletvisionlanguage} and Insight-V~\citep{dong2025insightvexploringlongchainvisual}, which demonstrated preliminary reasoning capabilities. This evolution reached new heights in LLMs such as GPT-5.5~\citep{openai2026gpt55} and DeepSeek-V4~\citep{deepseek2026v4}, which demonstrated human-like stepwise reasoning in tasks such as logical question-answering. The corresponding optimization strategies have given rise to LVLMs with reasoning capabilities, such as Qwen3.5~\citep{qwen3.5}.
To match the rapidly growing reasoning capability, stronger evaluation protocols are required to better evaluate the performance of LVLMs.

\begin{figure*}[htbp!]
  \centering
  \includegraphics[width=\textwidth]{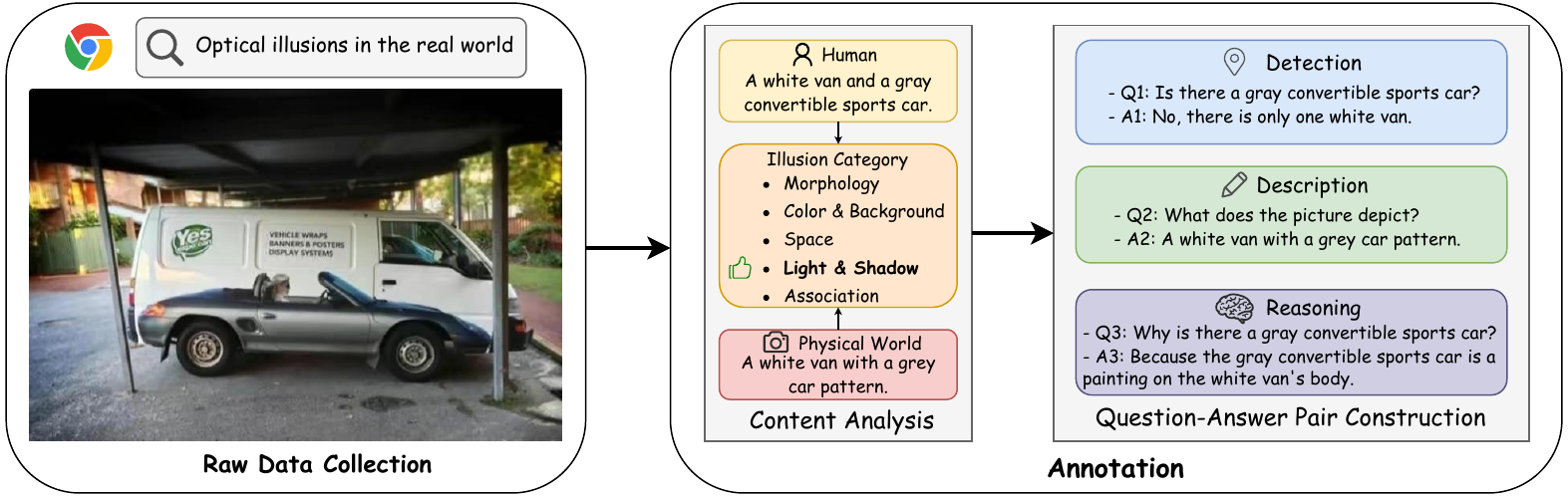}
  \caption{Overview of \textbf{\benchmarkname~}Construction. Images in \benchmarkname~are collected from search engines to ensure authenticity. The annotation consists of two stages: (1) Content Analysis, which involves the determination of illusion types by comparing human descriptions with ground-truth reality, and (2) QA Construction, which encompasses the creation of high-quality detection, description, and reasoning questions paired with answers.}
  \label{fig:annote_pipeline}
\end{figure*}

\subsection{Why Use Visual Illusions?}
LVLMs demonstrate solid performance in fundamental tasks such as visual question answering and image captioning. These tasks struggle to reflect the performance differences between different LVLMs. In contrast, visual illusion tasks, which are more complex, can effectively reveal the performance differences between various LVLMs.
To answer questions about illusions, LVLMs must conduct complex, multi-grained reasoning. This process requires comprehending the macroscopic semantic context while precisely pinpointing illusion-inducing features and integrating microscopic visual cues to see through the deceptive appearance and discern the underlying reality for the correct answer. 
Therefore, evaluation based on visual illusions can reflect the human-like degree of perception and reasoning capabilities, thereby offering a novel dimension beyond general evaluation.


\section{\benchmarkname~Construction} \label{category}
Visual illusion is defined as a phenomenon in which the perception of an observer regarding a physical attribute, including length, angle, color, or object motion, systematically deviates from objective reality under specific visual conditions. Existing evaluations primarily focus on the perceptual capabilities of LVLMs, while assessments of their reasoning abilities have largely relied on non-natural images.

To bridge this gap, we constructed a new benchmark, \textbf{\benchmarkname}. Employing natural illusion images, we aim to comprehensively evaluate both the perceptual and reasoning capabilities of LVLMs. The overall construction pipeline of \benchmarkname~is illustrated in Figure~\ref{fig:annote_pipeline}.

\subsection{Key Objectives}
A rigorous evaluation of the capabilities of LVLMs in handling illusions should follow these objectives:
\begin{itemize}[leftmargin=*]
    \item \textit{Minimizing prior knowledge bias with real-world scenarios}. To avoid the memorization interference caused by canonical synthetic illusions ingested during training, evaluations must prioritize authentic, real-world images. This ensures a measurement of genuine visual perception rather than mere training data memorization.
    \item \textit{Integrating binary and image-specific open-ended QA}. Relying exclusively on binary settings allows models to exploit random guessing without true comprehension. Combining binary questions with open-ended inquiries that are carefully tailored to specific visual details enables a comprehensive assessment of visual understanding across distinct levels of task complexity.
\end{itemize}

\subsection{Taxonomy of Visual Illusions}
Based on the underlying causes, we classify the visual illusions in our benchmark into five distinct categories:
\begin{itemize}[leftmargin=*]
    \item \textit{Morphological Illusion}: Visual illusions caused by shape or occlusion. They mislead observers into misperceiving the integrity of a single object.
    
    \item \textit{Color and Background Illusion}: Visual illusions caused by the color, brightness, texture, and contrast of an object against the background. These either cause incorrect perceptions of the object itself or lead to fusion with the background.
    
    \item \textit{Spatial Illusion}: Visual illusions caused by perspective, distance, arrangement, or occlusion between objects, which can lead to misperceptions of size, position, depth, or relative relationships.
    
    \item \textit{Light and Shadow Illusion}: Visual illusions caused by the effects of light, shadow, reflection, or refraction, distorting perceptions of the shape, brightness, or color of an object.
    
    \item \textit{Associative Illusion}: Visual illusions caused by the shape, texture, or arrangement of an object resembling something familiar in human memory, triggering semantic associations. This misleads observers into incorrectly recognizing the object.
\end{itemize}


\subsection{Data Collection}
The images in \benchmarkname~are collected from the real world. By querying search engines with the keyword ``optical illusions in the real world'', we can obtain a vast number of illusion images. We manually verified and filtered these images one by one. Images were excluded if they were blurry, irrelevant to visual illusions, or subject to copyright restrictions. 
Ultimately, we obtained a collection of authentic, high-quality illusion images.
Although it is possible to generate hallucinatory-like images through data synthesis, such images exhibit partial visual logical errors, providing shortcuts for LVLMs when generating responses. Using visual illusion images from the real world ensures that \benchmarkname~is challenging.

\subsection{Annotation} \label{question_level}

We design a three-stage pipeline to process the collected images alongside metadata detailing the initial misperception of the uploader and the actual physical content. First, three independent annotators report their initial visual impressions; we retain only images aligning with the documented misperception. They also verify the factual accuracy of the metadata. This verification establishes validated pairs contrasting physical reality and human misperception. Second, annotators independently classify the cause of the illusion into five categories, resolving disagreements via majority voting. Finally, a fourth annotator reviews the finalized data for formatting and linguistic quality.


For these images, we construct three types of questions: \textit{\textbf{Detection}}, \textit{\textbf{Description}}, and \textit{\textbf{Reasoning}}. Detection questions are binary inquiries verifying object existence or attribute correctness. Description questions are open-ended inquiries to target specific attributes or request general content summaries, determining whether models exhibit visual illusions similar to humans. Reasoning questions evaluate whether LVLMs can infer the causes of the illusion by understanding the discrepancy between physical reality and human misperception.

Annotators formulate QA pairs focusing on illusion-related regions, maximizing diversity and quantity without compromising accuracy. Each pair then undergoes at least two rounds of cross-verification and revision by independent annotators. This process filters out ambiguous entries, ensuring every question has a definitive and correct answer.

\subsection{Statistics}

\begin{figure}[htbp!]
  \centering
  \includegraphics[width=0.7\linewidth]{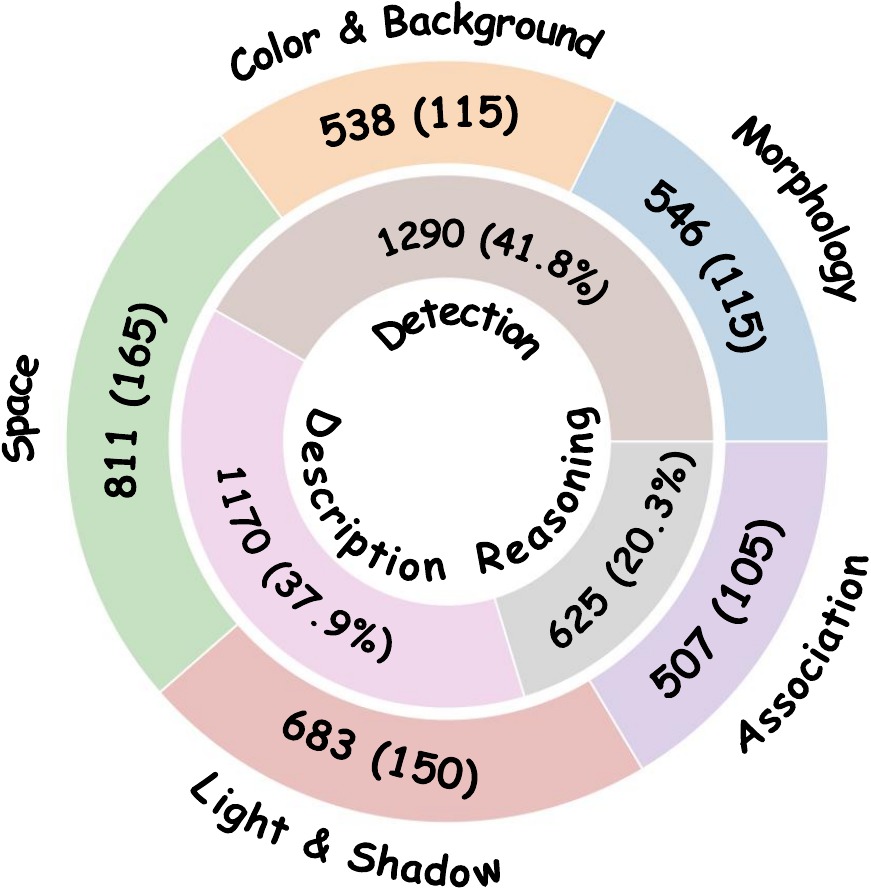}
  \caption{Statistics of \benchmarkname. The inner ring shows question categories, with numbers representing the count and proportion of questions. The outer ring shows illusion categories, with numbers representing the count of questions and images.}
  \label{fig:benchmark_part}
\end{figure}


As shown in Figure~\ref{fig:benchmark_part}, \benchmarkname~consists of 650 unedited real-world illusion images across five categories, expanding upon the Real-scene subset of IllusionVQA (60 images). To ensure novelty, image duplication rates with IllusionVQA and IllusionBench+ are strictly kept below 3\% and 10\%, respectively. We construct over 3,000 QA pairs (far exceeding the 435 pairs of IllusionVQA), distributed across detection, description, and reasoning tasks at an approximate 2:2:1 ratio. Binary questions are limited to mitigate random guessing. Overall, \benchmarkname~comprehensively evaluates and distinguishes the perception and reasoning capabilities of various LVLMs.

\subsection{Comparison with Existing Benchmarks}
\begin{figure}[htbp!]
  \centering
  \includegraphics[width=0.9\linewidth]{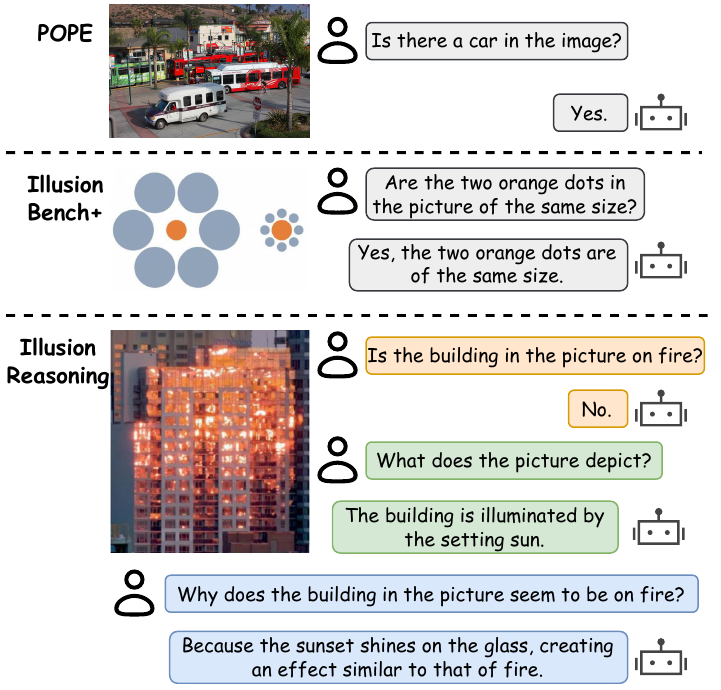}
  \caption{Comparison between \benchmarkname~and existing visual understanding benchmarks.}
  \label{fig:comparison_benchmarks}
\end{figure}

Although there are already some benchmarks about hallucination and illusion, \benchmarkname~is significantly different from them. We present examples from various benchmarks in Figure~\ref{fig:comparison_benchmarks} to illustrate the differences.

\begin{table*}[t]
\centering
\resizebox{\textwidth}{!}
{
\begin{tabular}{llcccccc} \hline
\multicolumn{1}{l}{\multirow{2}{*}{\textbf{\textit{Model}}}} & \multicolumn{1}{l}{\multirow{2}{*}{\textbf{\textit{Size}}}} & \multicolumn{5}{c}{\textbf{\textit{\benchmarkname}}}          & \multicolumn{1}{c}{\multirow{2}{*}{\textbf{\textit{Avg}}}}                                 \\ \cline{3-7} 
\multicolumn{1}{c}{}                       & \multicolumn{1}{c}{}                      & Morphology & \makecell[r]{Color \& \\ Background}  & Space & \makecell[c]{Light \& \\ Shadow} & Association & \\ \hline
\multicolumn{8}{c}{\cellcolor[HTML]{EFEFEF}\textbf{\textit{Closed-Source}}}                                                                              \\ \hline
                Gemini-3.1-Pro    & -             &\textbf{57.14}        &\textbf{57.81}    &\underline{58.57}   &\textbf{66.18}            &\underline{67.85}       &\textbf{61.39}  \\
                GPT-5.5       & -             &50.73        &55.39    &\textbf{60.69}   &\underline{59.88}            &\textbf{69.03}       &\underline{59.22}  \\
                Seed-2.0        & -             &\underline{52.93}        &\underline{55.76}    &57.46   &51.68           &64.69           &56.27  \\  \hline
\multicolumn{8}{c}{\cellcolor[HTML]{EFEFEF}\textbf{\textit{Open-Source}}}                           \\ 

                Qwen3.5       &397B(A17B)     &55.49       &52.60    &58.08   &57.98            &61.93            &57.28 \\
                InternVL3.5   &241B(A28B)     &48.35             &\underline{48.88}        &45.48        &47.73               &48.13            &47.49 \\ 
                GLM-4.6V       &106B(A12B)     &43.59        &47.21    &41.55  &49.78           &41.81             &44.76 \\
                InternVL3.5   &38B            &41.21            &44.24         &34.65        &44.07        &43.79            &41.07 \\ 
                Qwen3.6        &35B(A3B)       &46.89        &47.21    &57.83   &54.32            &56.41       &52.93 \\ 
                Qwen3.6       &27B            &52.38        &54.28    &56.72  &51.83            &54.24       &53.71\\           
                InternVL3.5    &14B            &34.25          &36.80         &31.69        &35.72                 &41.42            &35.53 \\ 
                GLM-4.6V      &9B             & 48.72         &51.86         &34.77        &44.66                 &42.01            &43.60 \\            
                Qwen3.5       &9B             &44.14         &45.54         &53.88        &47.00                 &47.73            &48.17  \\
                LLaVA-1.6       &7B           &25.46            &42.01         &25.40        &39.24           &36.88            &33.26 \\ 
                MiMo-VL        &7B             &49.63           &50.74         &32.80        &50.07                 & 44.97           &44.73 \\ 
                Phi-4 Multimodal          &6B    &18.42          &30.56             &19.83         &23.30        &29.20                 & 24.19 \\ 
                Qwen3.5       &4B               &52.36            &49.44         &52.65        &54.47                &46.55            &51.47 \\ 
                InternVL3.5    &2B               &28.39             &32.16        &27.00       &31.92                &37.87            &31.02 \\ 
                Qwen3.5       &2B               &29.30           & 37.73        &33.79        &44.36                 &42.41            &37.43\\ 
                InternVL3.5   &1B               &25.09            &30.48         &24.04        &26.65                 &32.15           &27.26 \\ \hline
\end{tabular}
}
\caption{\textbf{Results on \benchmarkname.} We report accuracy judged by GPT-4o. Best results are shown in \textbf{bold}. The suboptimal results are indicated by \underline{underline}.}
\label{tab:overall_results}
\end{table*}

\benchmarkname~demonstrates several advantages: (1) It covers perception and reasoning tasks. Compared to previous perception-focused benchmarks, it comprehensively evaluates the capabilities of LVLMs.
(2) \benchmarkname~evaluates whether LVLMs can accurately understand visual information, while hallucination benchmarks like POPE~\citep{li2023evaluating} focus on evaluating the factual accuracy of the content generated by LVLMs.
(3) Compared to the many synthetic images in the IllusionBench+~\citep{zhang2025illusionbenchlargescalecomprehensivebenchmark}, the images in \benchmarkname~are sourced from the real world and maintain authenticity.
\section{Experiments}
\subsection{Choices of LVLMs}
We evaluate a diverse set of LVLMs, spanning both closed-source and open-source variants, with parameter counts ranging from several billions to hundreds of billions.
For LVLMs equipped with reasoning capabilities, we primarily compare their performance in the instruction (non-thinking) mode, as shown in Table~\ref{tab:overall_results}. In subsequent sections, we further analyzed the performance of these LVLMs in both thinking and non-thinking modes on \benchmarkname.

\subsection{Evaluation Setting}
To ensure a fair comparison, all LVLMs receive identical single-turn prompts consisting solely of an image and a query, 
with no prior conversational context. For LVLMs like LLaVA-1.6~\citep{llava-next}, the input is organized as follows using the template: ``\texttt{USER: <|image|> \{question\} ASSISTANT:}''. To facilitate automated evaluation, we explicitly append formatting instructions (e.g., ``\textit{Provide an explicit conclusion.}''), ensuring that outputs are structured and easy to judge.

\subsection{LLM as a Judge}
In addition to serving as answer generators, LLMs offer a compelling alternative to traditional expert-driven evaluation~\citep{gu2025surveyllmasajudge}. Beyond binary questions, \benchmarkname~incorporates free-form questions whose answers do not conform to a predefined structure. In order to conduct a precise evaluation, we employ GPT-4o as the evaluator to determine whether a model’s response aligns with the annotated answers, inspired by OpenCompass~\citep{2023opencompass}. We randomly sampled 200 cases and evaluated the differences between GPT-4o and human evaluation, finding that it achieves a consistency rate of 99\%. For the parts where there are inconsistencies, we provide detailed cases in the Appendix~\ref{inconsistance} to illustrate the reasons.


\section{Main Results}

\subsection{Overall Performance}

Our evaluation covers a range of widely used LVLMs. And their performance on ~\benchmarkname~are summarized in Table~\ref{tab:overall_results}. 
As indicated in the table, neither closed-source nor open-source LVLMs achieved satisfactory results compared to other benchmarks such as POPE~\citep{li2023evaluating} and IllusionVQA~\citep{illusionvqa}, demonstrating the challenge of \benchmarkname. Evaluations based on \benchmarkname~provide more discriminative insights into LVLMs' capabilities. Open-source LVLMs underperformed closed-source ones on \benchmarkname. However, the gap is not insurmountable, and their performance is comparable in some specialized sub-tasks. 
Among the five illusion categories in \benchmarkname, LVLMs performed worst on morphology-related illusions and best on association-related ones.  

Results from open-source models of various sizes, ranging from 1B to hundreds of billions, indicate that the performance of LVLMs on \benchmarkname~is not determined by parameter size: small LVLMs can match the performance of models several times their scale. 
We also found that the performance is irrelevant to model architecture, with Qwen3.6 delivering similar performance across different model architectures. For example, Qwen3.6 shows similar performance across different architectures: MOE and Dense. This indicates that the performance of LVLMs on illusion mainly depends on the data.

\subsection{Case Study}
\begin{figure}[htbp!]
  \centering
  \includegraphics[width=\linewidth]{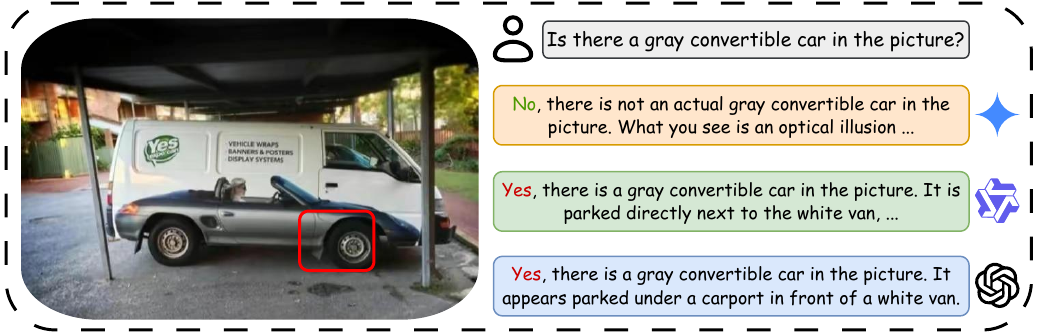}
  \caption{Result comparison between Gemini-3.1-Pro, Qwen3.5-397B, and GPT-5.5. The characteristics within the \textcolor{red}{red box} indicate that the gray car does not actually exist.}
  \label{fig:case_study}
\end{figure}

To provide an intuitive analysis of the performance among LVLMs in \benchmarkname~, we demonstrated the different responses produced by LVLMs when given the same input in Figure~\ref{fig:case_study}. The example we chose is an image of a van with the gray convertible spray-painted on its right side. We feed this image to LVLMs, asking them to judge whether a real gray convertible car exists in the scene.
We found that even the most advanced LVLMs struggle to provide the correct answer. 
This demonstrates that there is still considerable space for the reasoning capabilities of LVLMs to improve. Visual illusions can be used as a challenging task to measure the capability of LVLMs.

\section{Further Analysis}
\subsection{Can thinking help LVLMs to recognize illusions?}

Recent LVLMs have integrated reasoning capabilities through introducing thinking mode, demonstrating performance improvements on different evaluations. Whether thinking is effective for illusion recognition is to be examined.
We conducted preliminary attempts based on \benchmarkname. And the results are shown in Table~\ref{tab:thinking_mode}. We evaluated LVLMs across different question categories: detection, description, and reason in both thinking and non-thinking modes. 

\begin{table}[htbp!]
\centering
\resizebox{0.9\linewidth}{!}{
\begin{tabular}{llrrr} \hline
 & \textbf{\textit{Size}} & \textbf{\textit{Detection}} & \textbf{\textit{Description}} & \textbf{\textit{Reason}}  \\ \hline
Gemini-3.1-Pro &-      &65.27         &54.87         &65.60     \\
\rowcolor{gray!20} Gemini-3.1-Pro &-     &67.83         &56.32         &\textbf{68.96}          \\
GPT-5.5 &-     &64.49         &51.54        &62.72              \\
\rowcolor{gray!20} GPT-5.5 &-      &\textbf{68.22}         &54.27         &\underline{66.88}             \\
Seed-2.0 &-      &58.53        &\underline{56.67}  &50.88              \\
\rowcolor{gray!20} Seed-2.0 &-      &62.87         &\textbf{59.32} & 60.32              \\

Qwen3.5 &397B(A17B)      &\underline{67.98}         &45.47         &57.28              \\ 
\rowcolor{gray!20} Qwen3.5 &397B(A17B) &64.65         &48.03         &61.76 \\ 

GLM-4.6V &106B(A12B)      &47.60        &43.93         &40.48              \\
\rowcolor{gray!20} GLM-4.6V &106B(A12B)      &52.48        &47.69         &44.64\\

Qwen3.6 &35B(A3B)     &62.79        &43.59         &50.08 \\
\rowcolor{gray!20}Qwen3.6 &35B(A3B)             &59.15        &45.90         &59.20 \\

Qwen3.6 &27B              &62.09        &44.36         &53.92  \\
\rowcolor{gray!20}Qwen3.6 &27B               &60.85      &46.24         &62.24 \\\hline
\end{tabular}}
\caption{Performance comparison of LVLMs in thinking and non-thinking modes. Optimal results are shown in \textbf{bold}, and suboptimal results are denoted by \underline{underlining}. \colorbox{gray!20}{Gray} indicates responses yielded in thinking mode.}
\label{tab:thinking_mode}
\end{table}

Empirical results indicate that for tasks requiring reasoning about the underlying mechanisms of visual illusions, LVLMs exhibit a significant performance disparity between the thinking and non-thinking modes. However, in detection and description questions that focused on perceptual evaluation, LVLMs exhibit unstable performance improvements in thinking mode, even showing a declining trend, such as Qwen3.5 and Qwen3.6. This suggests that the integration of the perception and reasoning capabilities of LVLMs needs to be improved further.


\subsection{Are LVLMs aligned with human intuition?}

\begin{figure}[htbp!]
  \centering
  \includegraphics[width=\linewidth]{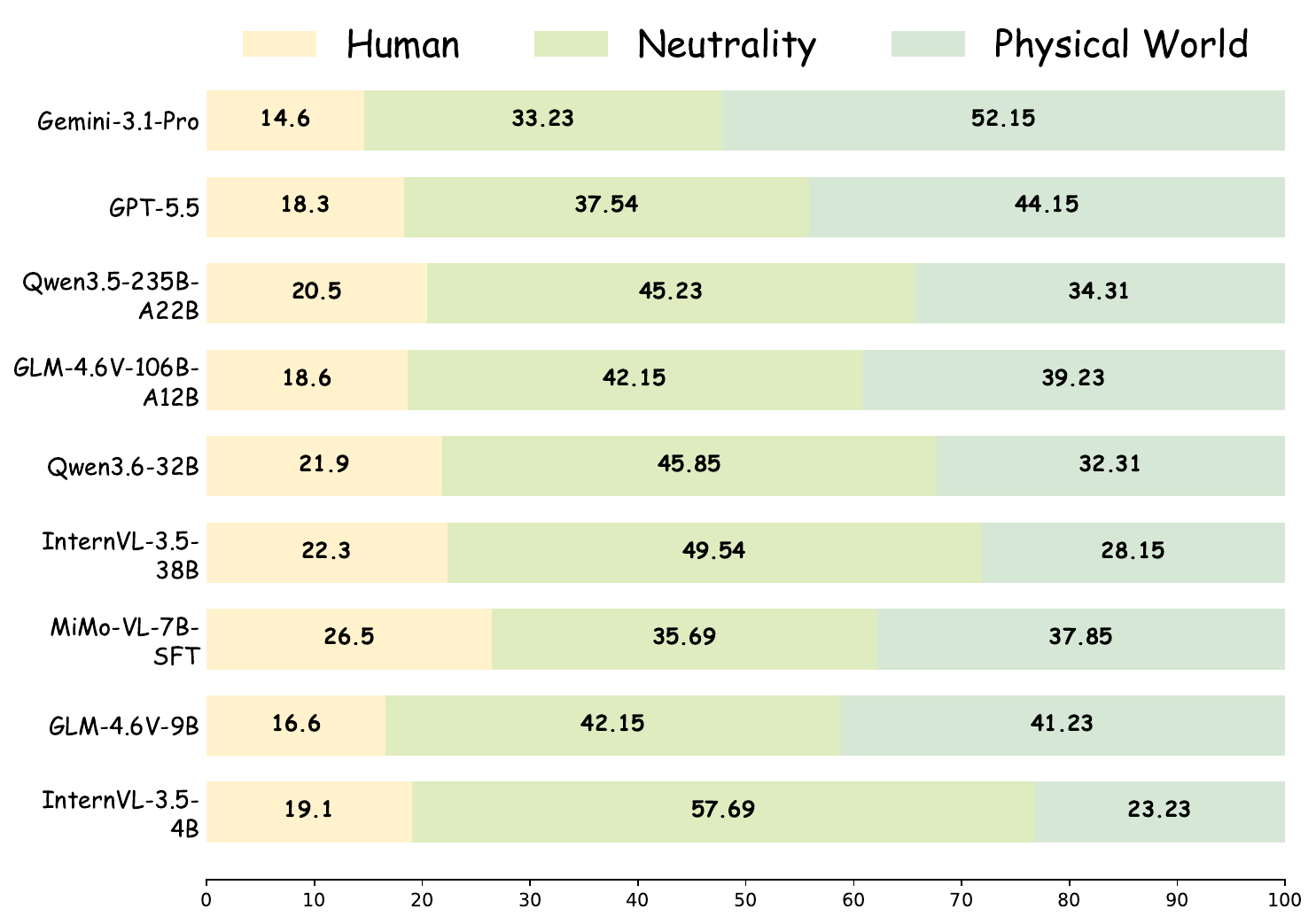}
  \caption{Comparison of LVLMs' preference for alignment between the human intuition and physical world.}
  \label{fig:alignment_ablation}
\end{figure}

We have demonstrated in Table~\ref{tab:overall_results} that the performance of existing LVLMs on {\benchmarkname} is not satisfactory. We would like to analyze whether this should be attributed to their alignment with human intuition, i.e., LVLMs make mistakes like humans.
We instruct LVLMs to describe the image without other restrictions and analyze the preference of responses. Besides preference to \textit{Human} (illusion) and \textit{physical world} (truth), we set a third category \textit{Neutrality} for those responses lacking clear preference. Results are shown in Figure~\ref{fig:alignment_ablation}.


While some responses are aligned with human intuition, there are large portions of responses that fall into the \textit{Neutrality} category, ranging approximately from 30\% to 60\% across different models. Most of such responses share a common feature: they simply ignore the core image content regarding the illusion, and make safe claims on less important content. A corresponding example is shown in Appendix~\ref{alignment}.

Such observation points out the direction for future optimisation. Safe responses that avoid describing ambiguous content may be regarded as correct during alignment optimisation, if the judging criterion is simply about making no mistakes. However, safe responses miss important information and should not be preferred. To discourage the generation of such responses, we suggest refined strategies for alignment optimisation. For example, one could force the model to answer various questions regarding the core content, so as to avoid shortcuts for scoring.



\subsection{Can LVLMs identify the underlying cause and correct logical errors?}
In this part, we analyzed whether LVLMs can provide correct reasons when being aware of the differences between real-world and illusory content. 
Specifically, we employed two distinct task formats: multiple-choice and free-form QA to evaluate the reasoning capabilities of LVLMs. For an image, a physical-world description, and a human-annotated illusion, the multi-choice task prompts LVLMs to select two relevant categories from the five illusion categories listed in Section~\ref{category}. And the free-form QA tasks need LVLMs to provide a reason given physical reality and illusion. We evaluated whether the types of illusions annotated appeared in the output yielded by LVLMs. Results are displayed in the Table~\ref{tab:reasoning_ablation}.

\begin{table}[htbp!]
\centering

\resizebox{0.9\linewidth}{!}{
\begin{tabular}{llrrr} \hline
 & \textbf{\textit{Size}} & \textbf{\textit{Choice}} & \textbf{\textit{Free-form}}   \\ \hline
Gemini-3.1-Pro &-          &73.50         &65.60              \\
\rowcolor{gray!20}Gemini-3.1-Pro &-     &\textbf{80.83}         &\textbf{68.96}                 \\
GPT-5.5 &-      &75.17         &62.72                    \\
\rowcolor{gray!20}GPT-5.5 &-      &\underline{77.83}         &\underline{66.88}                   \\
Seed-2.0 &-      &58.50        &50.88               \\
\rowcolor{gray!20}Seed-2.0 &-      &64.17         &60.32              \\
GLM-4.6V &106B(A12B)      &55.00          &40.48                    \\ 
\rowcolor{gray!20}GLM-4.6V &106B(A12B)      &49.33        &44.64                     \\
Qwen3.5 &397B(A17B)      &69.00          &57.28                     \\ 
\rowcolor{gray!20} Qwen3.5 &397B(A17B)      &66.83        &61.76                      \\  
Qwen3.6 &35B(A3B)           &73.17        &50.08          \\
\rowcolor{gray!20}Qwen3.6 &35B(A3B)            &67.67       &59.20         \\
Qwen3.6 &27B             &64.67        &53.92          \\
\rowcolor{gray!20}Qwen3.6 &27B             &59.50       &62.24          \\ \hline
\end{tabular}}
\caption{Comparison of LVLMs' ability to reason about visual illusions in \colorbox{gray!20}{thinking} and non-thinking modes.}
\label{tab:reasoning_ablation}
\end{table}

We found that under different modes of inference, the performances of the two tasks exhibited distinct trends. In the free-form task, LVLMs can provide precise explanations for the formation of illusions through thinking. 
When we provide the category of illusions generated and formalize the question of their causes as multi-choice tasks, LVLMs actually perform better without engaging in thinking.
This indicates that while LVLMs incorporate reasoning capabilities, they also carry the risk of overthinking, which may compromise performance on simple reasoning tasks. Due to space constraints, we provided a detailed example in Appendix~\ref{overthinking} to illustrate this issue.

\section{Conclusion}
In this work, we analyze the reasoning capabilities of existing LVLMs using visual illusions. We collected a set of real-world illusion images and constructed question-answer pairs designed to evaluate the perceptual and reasoning capabilities of LVLMs, forming a new benchmark: \benchmarkname. Evaluation based on \benchmarkname~for open-source and closed-source LVLMs indicates that the reasoning capabilities of these models are not as competitive as claimed. And not all questions benefit from reasoning; the integration of LVLMs’ perceptual and reasoning capabilities requires further exploration. Further analysis of \benchmarkname~ reveals the problem in optimization, emphasising the alignment of content while neglecting the alignment of focus. We hope that \benchmarkname~can be widely used by the community as an insightful benchmark.

\section*{Limitations}
Illusions, as phenomena where human intuition diverges from reality, create an inherent conflict between perception and reality. We constructed \benchmarkname, a benchmark comprising illusion images to evaluate the perceptual and reasoning capabilities of LVLMs. However, constrained by the stringent conditions under which illusions occur in the physical world, the volume of data in \benchmarkname~is insufficient to support training. We will continue to collect similar images and consider how to incorporate illusion images into the training process.

\section*{Acknowledgments}
This work is supported in part by the National Natural Science Foundation of China (NO.62676019, No.62106259), the Youth Program of the National Natural Science Foundation of China (Grant No.62406309), and the Hebei Provincial Department of Science And Technology Foundation under Grant (No.26240601D).

\bibliography{custom}

@misc{meta2025llama4,
    author = {Meta-AI},
    title = {The Llama 4 herd: The beginning of a new era of natively multimodal AI innovation},
    year = {2025},
    url = {https://ai.meta.com/blog/llama-4-multimodal-intelligence/}
}

@misc{liu2024mmbenchmultimodalmodelallaround,
      title={MMBench: Is Your Multi-modal Model an All-around Player?}, 
      author={Yuan Liu and Haodong Duan and Yuanhan Zhang and Bo Li and Songyang Zhang and Wangbo Zhao and Yike Yuan and Jiaqi Wang and Conghui He and Ziwei Liu and Kai Chen and Dahua Lin},
      year={2024},
      eprint={2307.06281},
      archivePrefix={arXiv},
      primaryClass={cs.CV},
      url={https://arxiv.org/abs/2307.06281}, 
}

@misc{mmstar,
      title={Are We on the Right Way for Evaluating Large Vision-Language Models?}, 
      author={Lin Chen and Jinsong Li and Xiaoyi Dong and Pan Zhang and Yuhang Zang and Zehui Chen and Haodong Duan and Jiaqi Wang and Yu Qiao and Dahua Lin and Feng Zhao},
      year={2024},
      eprint={2403.20330},
      archivePrefix={arXiv},
      primaryClass={cs.CV},
      url={https://arxiv.org/abs/2403.20330}, 
}

@misc{simplevqa,
      title={SimpleVQA: Multimodal Factuality Evaluation for Multimodal Large Language Models}, 
      author={Xianfu Cheng and Wei Zhang and Shiwei Zhang and Jian Yang and Xiangyuan Guan and Xianjie Wu and Xiang Li and Ge Zhang and Jiaheng Liu and Yuying Mai and Yutao Zeng and Zhoufutu Wen and Ke Jin and Baorui Wang and Weixiao Zhou and Yunhong Lu and Tongliang Li and Wenhao Huang and Zhoujun Li},
      year={2025},
      eprint={2502.13059},
      archivePrefix={arXiv},
      primaryClass={cs.CL},
      url={https://arxiv.org/abs/2502.13059}, 
}

@misc{openai2024gpt4technicalreport,
      title={GPT-4 Technical Report}, 
      author={OpenAI and Josh Achiam and Steven Adler and Sandhini Agarwal and Lama Ahmad and Ilge Akkaya and Florencia Leoni Aleman and Diogo Almeida and Janko Altenschmidt and Sam Altman and Shyamal Anadkat and Red Avila and Igor Babuschkin and Suchir Balaji and Valerie Balcom and Paul Baltescu and Haiming Bao and Mohammad Bavarian and Jeff Belgum and Irwan Bello and Jake Berdine and Gabriel Bernadett-Shapiro and Christopher Berner and Lenny Bogdonoff and Oleg Boiko and Madelaine Boyd and Anna-Luisa Brakman and Greg Brockman and Tim Brooks and Miles Brundage and Kevin Button and Trevor Cai and Rosie Campbell and Andrew Cann and Brittany Carey and Chelsea Carlson and Rory Carmichael and Brooke Chan and Che Chang and Fotis Chantzis and Derek Chen and Sully Chen and Ruby Chen and Jason Chen and Mark Chen and Ben Chess and Chester Cho and Casey Chu and Hyung Won Chung and Dave Cummings and Jeremiah Currier and Yunxing Dai and Cory Decareaux and Thomas Degry and Noah Deutsch and Damien Deville and Arka Dhar and David Dohan and Steve Dowling and Sheila Dunning and Adrien Ecoffet and Atty Eleti and Tyna Eloundou and David Farhi and Liam Fedus and Niko Felix and Simón Posada Fishman and Juston Forte and Isabella Fulford and Leo Gao and Elie Georges and Christian Gibson and Vik Goel and Tarun Gogineni and Gabriel Goh and Rapha Gontijo-Lopes and Jonathan Gordon and Morgan Grafstein and Scott Gray and Ryan Greene and Joshua Gross and Shixiang Shane Gu and Yufei Guo and Chris Hallacy and Jesse Han and Jeff Harris and Yuchen He and Mike Heaton and Johannes Heidecke and Chris Hesse and Alan Hickey and Wade Hickey and Peter Hoeschele and Brandon Houghton and Kenny Hsu and Shengli Hu and Xin Hu and Joost Huizinga and Shantanu Jain and Shawn Jain and Joanne Jang and Angela Jiang and Roger Jiang and Haozhun Jin and Denny Jin and Shino Jomoto and Billie Jonn and Heewoo Jun and Tomer Kaftan and Łukasz Kaiser and Ali Kamali and Ingmar Kanitscheider and Nitish Shirish Keskar and Tabarak Khan and Logan Kilpatrick and Jong Wook Kim and Christina Kim and Yongjik Kim and Jan Hendrik Kirchner and Jamie Kiros and Matt Knight and Daniel Kokotajlo and Łukasz Kondraciuk and Andrew Kondrich and Aris Konstantinidis and Kyle Kosic and Gretchen Krueger and Vishal Kuo and Michael Lampe and Ikai Lan and Teddy Lee and Jan Leike and Jade Leung and Daniel Levy and Chak Ming Li and Rachel Lim and Molly Lin and Stephanie Lin and Mateusz Litwin and Theresa Lopez and Ryan Lowe and Patricia Lue and Anna Makanju and Kim Malfacini and Sam Manning and Todor Markov and Yaniv Markovski and Bianca Martin and Katie Mayer and Andrew Mayne and Bob McGrew and Scott Mayer McKinney and Christine McLeavey and Paul McMillan and Jake McNeil and David Medina and Aalok Mehta and Jacob Menick and Luke Metz and Andrey Mishchenko and Pamela Mishkin and Vinnie Monaco and Evan Morikawa and Daniel Mossing and Tong Mu and Mira Murati and Oleg Murk and David Mély and Ashvin Nair and Reiichiro Nakano and Rajeev Nayak and Arvind Neelakantan and Richard Ngo and Hyeonwoo Noh and Long Ouyang and Cullen O'Keefe and Jakub Pachocki and Alex Paino and Joe Palermo and Ashley Pantuliano and Giambattista Parascandolo and Joel Parish and Emy Parparita and Alex Passos and Mikhail Pavlov and Andrew Peng and Adam Perelman and Filipe de Avila Belbute Peres and Michael Petrov and Henrique Ponde de Oliveira Pinto and Michael and Pokorny and Michelle Pokrass and Vitchyr H. Pong and Tolly Powell and Alethea Power and Boris Power and Elizabeth Proehl and Raul Puri and Alec Radford and Jack Rae and Aditya Ramesh and Cameron Raymond and Francis Real and Kendra Rimbach and Carl Ross and Bob Rotsted and Henri Roussez and Nick Ryder and Mario Saltarelli and Ted Sanders and Shibani Santurkar and Girish Sastry and Heather Schmidt and David Schnurr and John Schulman and Daniel Selsam and Kyla Sheppard and Toki Sherbakov and Jessica Shieh and Sarah Shoker and Pranav Shyam and Szymon Sidor and Eric Sigler and Maddie Simens and Jordan Sitkin and Katarina Slama and Ian Sohl and Benjamin Sokolowsky and Yang Song and Natalie Staudacher and Felipe Petroski Such and Natalie Summers and Ilya Sutskever and Jie Tang and Nikolas Tezak and Madeleine B. Thompson and Phil Tillet and Amin Tootoonchian and Elizabeth Tseng and Preston Tuggle and Nick Turley and Jerry Tworek and Juan Felipe Cerón Uribe and Andrea Vallone and Arun Vijayvergiya and Chelsea Voss and Carroll Wainwright and Justin Jay Wang and Alvin Wang and Ben Wang and Jonathan Ward and Jason Wei and CJ Weinmann and Akila Welihinda and Peter Welinder and Jiayi Weng and Lilian Weng and Matt Wiethoff and Dave Willner and Clemens Winter and Samuel Wolrich and Hannah Wong and Lauren Workman and Sherwin Wu and Jeff Wu and Michael Wu and Kai Xiao and Tao Xu and Sarah Yoo and Kevin Yu and Qiming Yuan and Wojciech Zaremba and Rowan Zellers and Chong Zhang and Marvin Zhang and Shengjia Zhao and Tianhao Zheng and Juntang Zhuang and William Zhuk and Barret Zoph},
      year={2024},
      eprint={2303.08774},
      archivePrefix={arXiv},
      primaryClass={cs.CL},
      url={https://arxiv.org/abs/2303.08774}, 
}

@misc{yang2025qwen3technicalreport,
      title={Qwen3 Technical Report}, 
      author={An Yang and Anfeng Li and Baosong Yang and Beichen Zhang and Binyuan Hui and Bo Zheng and Bowen Yu and Chang Gao and Chengen Huang and Chenxu Lv and Chujie Zheng and Dayiheng Liu and Fan Zhou and Fei Huang and Feng Hu and Hao Ge and Haoran Wei and Huan Lin and Jialong Tang and Jian Yang and Jianhong Tu and Jianwei Zhang and Jianxin Yang and Jiaxi Yang and Jing Zhou and Jingren Zhou and Junyang Lin and Kai Dang and Keqin Bao and Kexin Yang and Le Yu and Lianghao Deng and Mei Li and Mingfeng Xue and Mingze Li and Pei Zhang and Peng Wang and Qin Zhu and Rui Men and Ruize Gao and Shixuan Liu and Shuang Luo and Tianhao Li and Tianyi Tang and Wenbiao Yin and Xingzhang Ren and Xinyu Wang and Xinyu Zhang and Xuancheng Ren and Yang Fan and Yang Su and Yichang Zhang and Yinger Zhang and Yu Wan and Yuqiong Liu and Zekun Wang and Zeyu Cui and Zhenru Zhang and Zhipeng Zhou and Zihan Qiu},
      year={2025},
      eprint={2505.09388},
      archivePrefix={arXiv},
      primaryClass={cs.CL},
      url={https://arxiv.org/abs/2505.09388}, 
}

@misc{openai2024gpt4o,
    author = {OpenAI},
    title = {Hello GPT-4o},
    year = {2024},
    url = {https://openai.com/index/hello-gpt-4o/}
}

@misc{bai2025qwen3vltechnicalreport,
      title={Qwen3-VL Technical Report}, 
      author={Shuai Bai and Yuxuan Cai and Ruizhe Chen and Keqin Chen and Xionghui Chen and Zesen Cheng and Lianghao Deng and Wei Ding and Chang Gao and Chunjiang Ge and Wenbin Ge and Zhifang Guo and Qidong Huang and Jie Huang and Fei Huang and Binyuan Hui and Shutong Jiang and Zhaohai Li and Mingsheng Li and Mei Li and Kaixin Li and Zicheng Lin and Junyang Lin and Xuejing Liu and Jiawei Liu and Chenglong Liu and Yang Liu and Dayiheng Liu and Shixuan Liu and Dunjie Lu and Ruilin Luo and Chenxu Lv and Rui Men and Lingchen Meng and Xuancheng Ren and Xingzhang Ren and Sibo Song and Yuchong Sun and Jun Tang and Jianhong Tu and Jianqiang Wan and Peng Wang and Pengfei Wang and Qiuyue Wang and Yuxuan Wang and Tianbao Xie and Yiheng Xu and Haiyang Xu and Jin Xu and Zhibo Yang and Mingkun Yang and Jianxin Yang and An Yang and Bowen Yu and Fei Zhang and Hang Zhang and Xi Zhang and Bo Zheng and Humen Zhong and Jingren Zhou and Fan Zhou and Jing Zhou and Yuanzhi Zhu and Ke Zhu},
      year={2025},
      eprint={2511.21631},
      archivePrefix={arXiv},
      primaryClass={cs.CV},
      url={https://arxiv.org/abs/2511.21631}, 
}

@article{lu2024deepseek,
  title={Deepseek-vl: towards real-world vision-language understanding},
  author={Lu, Haoyu and Liu, Wen and Zhang, Bo and Wang, Bingxuan and Dong, Kai and Liu, Bo and Sun, Jingxiang and Ren, Tongzheng and Li, Zhuoshu and Yang, Hao and others},
  journal={arXiv preprint arXiv:2403.05525},
  year={2024}
}

@article{huang2024llm2clip,
  title={Llm2clip: Powerful language model unlocks richer visual representation},
  author={Huang, Weiquan and Wu, Aoqi and Yang, Yifan and Luo, Xufang and Yang, Yuqing and Hu, Liang and Dai, Qi and Wang, Chunyu and Dai, Xiyang and Chen, Dongdong and others},
  journal={arXiv preprint arXiv:2411.04997},
  year={2024}
}

@article{dosovitskiy2020image,
  title={An image is worth 16x16 words: Transformers for image recognition at scale},
  author={Dosovitskiy, Alexey},
  journal={arXiv preprint arXiv:2010.11929},
  year={2020}
}

@article{liu2023visual,
  title={Visual instruction tuning},
  author={Liu, Haotian and Li, Chunyuan and Wu, Qingyang and Lee, Yong Jae},
  journal={Advances in neural information processing systems},
  volume={36},
  pages={34892--34916},
  year={2023}
}

@inproceedings{chen2024internvl,
  title={Internvl: Scaling up vision foundation models and aligning for generic visual-linguistic tasks},
  author={Chen, Zhe and Wu, Jiannan and Wang, Wenhai and Su, Weijie and Chen, Guo and Xing, Sen and Zhong, Muyan and Zhang, Qinglong and Zhu, Xizhou and Lu, Lewei and others},
  booktitle={Proceedings of the IEEE/CVF conference on computer vision and pattern recognition},
  pages={24185--24198},
  year={2024}
}

@article{zhu2023minigpt,
  title={Minigpt-4: Enhancing vision-language understanding with advanced large language models},
  author={Zhu, Deyao and Chen, Jun and Shen, Xiaoqian and Li, Xiang and Elhoseiny, Mohamed},
  journal={arXiv preprint arXiv:2304.10592},
  year={2023}
}

@inproceedings{zhao2025beyond,
  title={Beyond sight: Towards cognitive alignment in lvlm via enriched visual knowledge},
  author={Zhao, Yaqi and Yin, Yuanyang and Li, Lin and Lin, Mingan and Huang, Victor Shea-Jay and Chen, Siwei and Chen, Weipeng and Yin, Baoqun and Zhou, Zenan and Zhang, Wentao},
  booktitle={Proceedings of the Computer Vision and Pattern Recognition Conference},
  pages={24950--24959},
  year={2025}
}

@article{li2023evaluating,
  title={Evaluating object hallucination in large vision-language models},
  author={Li, Yifan and Du, Yifan and Zhou, Kun and Wang, Jinpeng and Zhao, Wayne Xin and Wen, Ji-Rong},
  journal={arXiv preprint arXiv:2305.10355},
  year={2023}
}

@article{bai2024hallucination,
  title={Hallucination of multimodal large language models: A survey},
  author={Bai, Zechen and Wang, Pichao and Xiao, Tianjun and He, Tong and Han, Zongbo and Zhang, Zheng and Shou, Mike Zheng},
  journal={arXiv preprint arXiv:2404.18930},
  year={2024}
}

@misc{xu2025llavacotletvisionlanguage,
      title={LLaVA-CoT: Let Vision Language Models Reason Step-by-Step}, 
      author={Guowei Xu and Peng Jin and Ziang Wu and Hao Li and Yibing Song and Lichao Sun and Li Yuan},
      year={2025},
      eprint={2411.10440},
      archivePrefix={arXiv},
      primaryClass={cs.CV},
      url={https://arxiv.org/abs/2411.10440}, 
}

@misc{dong2025insightvexploringlongchainvisual,
      title={Insight-V: Exploring Long-Chain Visual Reasoning with Multimodal Large Language Models}, 
      author={Yuhao Dong and Zuyan Liu and Hai-Long Sun and Jingkang Yang and Winston Hu and Yongming Rao and Ziwei Liu},
      year={2025},
      eprint={2411.14432},
      archivePrefix={arXiv},
      primaryClass={cs.CV},
      url={https://arxiv.org/abs/2411.14432}, 
}

@misc{deepseekai2025deepseekr1incentivizingreasoningcapability,
      title={DeepSeek-R1: Incentivizing Reasoning Capability in LLMs via Reinforcement Learning}, 
      author={DeepSeek-AI and Daya Guo and Dejian Yang and Haowei Zhang and Junxiao Song and Ruoyu Zhang and Runxin Xu and Qihao Zhu and Shirong Ma and Peiyi Wang and Xiao Bi and Xiaokang Zhang and Xingkai Yu and Yu Wu and Z. F. Wu and Zhibin Gou and Zhihong Shao and Zhuoshu Li and Ziyi Gao and Aixin Liu and Bing Xue and Bingxuan Wang and Bochao Wu and Bei Feng and Chengda Lu and Chenggang Zhao and Chengqi Deng and Chenyu Zhang and Chong Ruan and Damai Dai and Deli Chen and Dongjie Ji and Erhang Li and Fangyun Lin and Fucong Dai and Fuli Luo and Guangbo Hao and Guanting Chen and Guowei Li and H. Zhang and Han Bao and Hanwei Xu and Haocheng Wang and Honghui Ding and Huajian Xin and Huazuo Gao and Hui Qu and Hui Li and Jianzhong Guo and Jiashi Li and Jiawei Wang and Jingchang Chen and Jingyang Yuan and Junjie Qiu and Junlong Li and J. L. Cai and Jiaqi Ni and Jian Liang and Jin Chen and Kai Dong and Kai Hu and Kaige Gao and Kang Guan and Kexin Huang and Kuai Yu and Lean Wang and Lecong Zhang and Liang Zhao and Litong Wang and Liyue Zhang and Lei Xu and Leyi Xia and Mingchuan Zhang and Minghua Zhang and Minghui Tang and Meng Li and Miaojun Wang and Mingming Li and Ning Tian and Panpan Huang and Peng Zhang and Qiancheng Wang and Qinyu Chen and Qiushi Du and Ruiqi Ge and Ruisong Zhang and Ruizhe Pan and Runji Wang and R. J. Chen and R. L. Jin and Ruyi Chen and Shanghao Lu and Shangyan Zhou and Shanhuang Chen and Shengfeng Ye and Shiyu Wang and Shuiping Yu and Shunfeng Zhou and Shuting Pan and S. S. Li and Shuang Zhou and Shaoqing Wu and Shengfeng Ye and Tao Yun and Tian Pei and Tianyu Sun and T. Wang and Wangding Zeng and Wanjia Zhao and Wen Liu and Wenfeng Liang and Wenjun Gao and Wenqin Yu and Wentao Zhang and W. L. Xiao and Wei An and Xiaodong Liu and Xiaohan Wang and Xiaokang Chen and Xiaotao Nie and Xin Cheng and Xin Liu and Xin Xie and Xingchao Liu and Xinyu Yang and Xinyuan Li and Xuecheng Su and Xuheng Lin and X. Q. Li and Xiangyue Jin and Xiaojin Shen and Xiaosha Chen and Xiaowen Sun and Xiaoxiang Wang and Xinnan Song and Xinyi Zhou and Xianzu Wang and Xinxia Shan and Y. K. Li and Y. Q. Wang and Y. X. Wei and Yang Zhang and Yanhong Xu and Yao Li and Yao Zhao and Yaofeng Sun and Yaohui Wang and Yi Yu and Yichao Zhang and Yifan Shi and Yiliang Xiong and Ying He and Yishi Piao and Yisong Wang and Yixuan Tan and Yiyang Ma and Yiyuan Liu and Yongqiang Guo and Yuan Ou and Yuduan Wang and Yue Gong and Yuheng Zou and Yujia He and Yunfan Xiong and Yuxiang Luo and Yuxiang You and Yuxuan Liu and Yuyang Zhou and Y. X. Zhu and Yanhong Xu and Yanping Huang and Yaohui Li and Yi Zheng and Yuchen Zhu and Yunxian Ma and Ying Tang and Yukun Zha and Yuting Yan and Z. Z. Ren and Zehui Ren and Zhangli Sha and Zhe Fu and Zhean Xu and Zhenda Xie and Zhengyan Zhang and Zhewen Hao and Zhicheng Ma and Zhigang Yan and Zhiyu Wu and Zihui Gu and Zijia Zhu and Zijun Liu and Zilin Li and Ziwei Xie and Ziyang Song and Zizheng Pan and Zhen Huang and Zhipeng Xu and Zhongyu Zhang and Zhen Zhang},
      year={2025},
      eprint={2501.12948},
      archivePrefix={arXiv},
      primaryClass={cs.CL},
      url={https://arxiv.org/abs/2501.12948}, 
}

@misc{openai2024openaio1card,
      title={OpenAI o1 System Card}, 
      author={OpenAI and : and Aaron Jaech and Adam Kalai and Adam Lerer and Adam Richardson and Ahmed El-Kishky and Aiden Low and Alec Helyar and Aleksander Madry and Alex Beutel and Alex Carney and Alex Iftimie and Alex Karpenko and Alex Tachard Passos and Alexander Neitz and Alexander Prokofiev and Alexander Wei and Allison Tam and Ally Bennett and Ananya Kumar and Andre Saraiva and Andrea Vallone and Andrew Duberstein and Andrew Kondrich and Andrey Mishchenko and Andy Applebaum and Angela Jiang and Ashvin Nair and Barret Zoph and Behrooz Ghorbani and Ben Rossen and Benjamin Sokolowsky and Boaz Barak and Bob McGrew and Borys Minaiev and Botao Hao and Bowen Baker and Brandon Houghton and Brandon McKinzie and Brydon Eastman and Camillo Lugaresi and Cary Bassin and Cary Hudson and Chak Ming Li and Charles de Bourcy and Chelsea Voss and Chen Shen and Chong Zhang and Chris Koch and Chris Orsinger and Christopher Hesse and Claudia Fischer and Clive Chan and Dan Roberts and Daniel Kappler and Daniel Levy and Daniel Selsam and David Dohan and David Farhi and David Mely and David Robinson and Dimitris Tsipras and Doug Li and Dragos Oprica and Eben Freeman and Eddie Zhang and Edmund Wong and Elizabeth Proehl and Enoch Cheung and Eric Mitchell and Eric Wallace and Erik Ritter and Evan Mays and Fan Wang and Felipe Petroski Such and Filippo Raso and Florencia Leoni and Foivos Tsimpourlas and Francis Song and Fred von Lohmann and Freddie Sulit and Geoff Salmon and Giambattista Parascandolo and Gildas Chabot and Grace Zhao and Greg Brockman and Guillaume Leclerc and Hadi Salman and Haiming Bao and Hao Sheng and Hart Andrin and Hessam Bagherinezhad and Hongyu Ren and Hunter Lightman and Hyung Won Chung and Ian Kivlichan and Ian O'Connell and Ian Osband and Ignasi Clavera Gilaberte and Ilge Akkaya and Ilya Kostrikov and Ilya Sutskever and Irina Kofman and Jakub Pachocki and James Lennon and Jason Wei and Jean Harb and Jerry Twore and Jiacheng Feng and Jiahui Yu and Jiayi Weng and Jie Tang and Jieqi Yu and Joaquin Quiñonero Candela and Joe Palermo and Joel Parish and Johannes Heidecke and John Hallman and John Rizzo and Jonathan Gordon and Jonathan Uesato and Jonathan Ward and Joost Huizinga and Julie Wang and Kai Chen and Kai Xiao and Karan Singhal and Karina Nguyen and Karl Cobbe and Katy Shi and Kayla Wood and Kendra Rimbach and Keren Gu-Lemberg and Kevin Liu and Kevin Lu and Kevin Stone and Kevin Yu and Lama Ahmad and Lauren Yang and Leo Liu and Leon Maksin and Leyton Ho and Liam Fedus and Lilian Weng and Linden Li and Lindsay McCallum and Lindsey Held and Lorenz Kuhn and Lukas Kondraciuk and Lukasz Kaiser and Luke Metz and Madelaine Boyd and Maja Trebacz and Manas Joglekar and Mark Chen and Marko Tintor and Mason Meyer and Matt Jones and Matt Kaufer and Max Schwarzer and Meghan Shah and Mehmet Yatbaz and Melody Y. Guan and Mengyuan Xu and Mengyuan Yan and Mia Glaese and Mianna Chen and Michael Lampe and Michael Malek and Michele Wang and Michelle Fradin and Mike McClay and Mikhail Pavlov and Miles Wang and Mingxuan Wang and Mira Murati and Mo Bavarian and Mostafa Rohaninejad and Nat McAleese and Neil Chowdhury and Neil Chowdhury and Nick Ryder and Nikolas Tezak and Noam Brown and Ofir Nachum and Oleg Boiko and Oleg Murk and Olivia Watkins and Patrick Chao and Paul Ashbourne and Pavel Izmailov and Peter Zhokhov and Rachel Dias and Rahul Arora and Randall Lin and Rapha Gontijo Lopes and Raz Gaon and Reah Miyara and Reimar Leike and Renny Hwang and Rhythm Garg and Robin Brown and Roshan James and Rui Shu and Ryan Cheu and Ryan Greene and Saachi Jain and Sam Altman and Sam Toizer and Sam Toyer and Samuel Miserendino and Sandhini Agarwal and Santiago Hernandez and Sasha Baker and Scott McKinney and Scottie Yan and Shengjia Zhao and Shengli Hu and Shibani Santurkar and Shraman Ray Chaudhuri and Shuyuan Zhang and Siyuan Fu and Spencer Papay and Steph Lin and Suchir Balaji and Suvansh Sanjeev and Szymon Sidor and Tal Broda and Aidan Clark and Tao Wang and Taylor Gordon and Ted Sanders and Tejal Patwardhan and Thibault Sottiaux and Thomas Degry and Thomas Dimson and Tianhao Zheng and Timur Garipov and Tom Stasi and Trapit Bansal and Trevor Creech and Troy Peterson and Tyna Eloundou and Valerie Qi and Vineet Kosaraju and Vinnie Monaco and Vitchyr Pong and Vlad Fomenko and Weiyi Zheng and Wenda Zhou and Wes McCabe and Wojciech Zaremba and Yann Dubois and Yinghai Lu and Yining Chen and Young Cha and Yu Bai and Yuchen He and Yuchen Zhang and Yunyun Wang and Zheng Shao and Zhuohan Li},
      year={2024},
      eprint={2412.16720},
      archivePrefix={arXiv},
      primaryClass={cs.AI},
      url={https://arxiv.org/abs/2412.16720}, 
}

@misc{liu2025phdchatgptpromptedvisualhallucination,
      title={PhD: A ChatGPT-Prompted Visual hallucination Evaluation Dataset}, 
      author={Jiazhen Liu and Yuhan Fu and Ruobing Xie and Runquan Xie and Xingwu Sun and Fengzong Lian and Zhanhui Kang and Xirong Li},
      year={2025},
      eprint={2403.11116},
      archivePrefix={arXiv},
      primaryClass={cs.CV},
      url={https://arxiv.org/abs/2403.11116}, 
}

@misc{gu2025surveyllmasajudge,
      title={A Survey on LLM-as-a-Judge}, 
      author={Jiawei Gu and Xuhui Jiang and Zhichao Shi and Hexiang Tan and Xuehao Zhai and Chengjin Xu and Wei Li and Yinghan Shen and Shengjie Ma and Honghao Liu and Saizhuo Wang and Kun Zhang and Yuanzhuo Wang and Wen Gao and Lionel Ni and Jian Guo},
      year={2025},
      eprint={2411.15594},
      archivePrefix={arXiv},
      primaryClass={cs.CL},
      url={https://arxiv.org/abs/2411.15594}, 
}

@article{illusionvqa,
  title={Illusionvqa: A challenging optical illusion dataset for vision language models},
  author={Shahgir, Haz Sameen and Sayeed, Khondker Salman and Bhattacharjee, Abhik and Ahmad, Wasi Uddin and Dong, Yue and Shahriyar, Rifat},
  journal={arXiv preprint arXiv:2403.15952},
  year={2024}
}

@misc{zhang2025illusionbenchlargescalecomprehensivebenchmark,
      title={IllusionBench+: A Large-scale and Comprehensive Benchmark for Visual Illusion Understanding in Vision-Language Models}, 
      author={Yiming Zhang and Zicheng Zhang and Xinyi Wei and Xiaohong Liu and Guangtao Zhai and Xiongkuo Min},
      year={2025},
      eprint={2501.00848},
      archivePrefix={arXiv},
      primaryClass={cs.CV},
      url={https://arxiv.org/abs/2501.00848}, 
}

@article{zhang2023grounding,
  title={Grounding visual illusions in language: Do vision-language models perceive illusions like humans?},
  author={Zhang, Yichi and Pan, Jiayi and Zhou, Yuchen and Pan, Rui and Chai, Joyce},
  journal={arXiv preprint arXiv:2311.00047},
  year={2023}
}

@inproceedings{guan2024hallusionbench,
  title={Hallusionbench: an advanced diagnostic suite for entangled language hallucination and visual illusion in large vision-language models},
  author={Guan, Tianrui and Liu, Fuxiao and Wu, Xiyang and Xian, Ruiqi and Li, Zongxia and Liu, Xiaoyu and Wang, Xijun and Chen, Lichang and Huang, Furong and Yacoob, Yaser and others},
  booktitle={Proceedings of the IEEE/CVF Conference on Computer Vision and Pattern Recognition},
  pages={14375--14385},
  year={2024}
}

@article{wang2023amber,
  title={Amber: An llm-free multi-dimensional benchmark for mllms hallucination evaluation},
  author={Wang, Junyang and Wang, Yuhang and Xu, Guohai and Zhang, Jing and Gu, Yukai and Jia, Haitao and Wang, Jiaqi and Xu, Haiyang and Yan, Ming and Zhang, Ji and others},
  journal={arXiv preprint arXiv:2311.07397},
  year={2023}
}

@misc{OlympiadBench,
      title={OlympiadBench: A Challenging Benchmark for Promoting AGI with Olympiad-Level Bilingual Multimodal Scientific Problems}, 
      author={Chaoqun He and Renjie Luo and Yuzhuo Bai and Shengding Hu and Zhen Leng Thai and Junhao Shen and Jinyi Hu and Xu Han and Yujie Huang and Yuxiang Zhang and Jie Liu and Lei Qi and Zhiyuan Liu and Maosong Sun},
      year={2024},
      eprint={2402.14008},
      archivePrefix={arXiv},
      primaryClass={cs.CL},
      url={https://arxiv.org/abs/2402.14008}, 
}

@misc{mathvista,
      title={MathVista: Evaluating Mathematical Reasoning of Foundation Models in Visual Contexts}, 
      author={Pan Lu and Hritik Bansal and Tony Xia and Jiacheng Liu and Chunyuan Li and Hannaneh Hajishirzi and Hao Cheng and Kai-Wei Chang and Michel Galley and Jianfeng Gao},
      year={2024},
      eprint={2310.02255},
      archivePrefix={arXiv},
      primaryClass={cs.CV},
      url={https://arxiv.org/abs/2310.02255}, 
}

@article{bach2006optical,
  title={Optical illusions},
  author={Bach, Michael and Poloschek, Charlotte M},
  journal={Adv Clin Neurosci Rehabil},
  volume={6},
  number={2},
  pages={20--21},
  year={2006}
}

@article{gregory1997knowledge,
  title={Knowledge in perception and illusion},
  author={Gregory, Richard L},
  journal={Philosophical Transactions of the Royal Society of London. Series B: Biological Sciences},
  volume={352},
  number={1358},
  pages={1121--1127},
  year={1997},
  publisher={The Royal Society}
}

@article{gregory1968perceptual,
  title={Perceptual illusions and brain models},
  author={Gregory, Richard Langton},
  journal={Proceedings of the Royal Society of London. Series B. Biological Sciences},
  volume={171},
  number={1024},
  pages={279--296},
  year={1968},
  publisher={The Royal Society London}
}

@article{sun2021imagenet,
  title={ImageNet-trained deep neural networks exhibit illusion-like response to the Scintillating grid},
  author={Sun, Eric D and Dekel, Ron},
  journal={Journal of Vision},
  volume={21},
  number={11},
  pages={15--15},
  year={2021},
  publisher={The Association for Research in Vision and Ophthalmology}
}

@article{garcia2015muller,
  title={The M{\"u}ller-Lyer illusion as seen by an artificial neural network},
  author={Garc{\'\i}a-Garibay, Otto B and de Lafuente, Victor},
  journal={Frontiers in computational neuroscience},
  volume={9},
  pages={21},
  year={2015},
  publisher={Frontiers Media SA}
}

@article{plotnitsky1997penrose,
  title={Penrose's triangles: The large, the small, and the human mind},
  author={Plotnitsky, Arkady},
  journal={Postmodern Culture},
  volume={7},
  number={3},
  year={1997},
  publisher={Johns Hopkins University Press}
}

@misc{openai2026gpt55,
  title        = {GPT-5.5 System Card},
  author       = {{OpenAI}},
  year         = {2026},
  month        = {April},
  howpublished = {\url{https://openai.com/index/gpt-5-5-system-card/}},
  note         = {Accessed: 2026-05-22}
}

@misc{deepseek2026v4,
  title        = {DeepSeek-V4 Technical Report},
  author       = {{DeepSeek-AI}},
  year         = {2026},
  month        = {April},
  howpublished = {\url{https://github.com/deepseek-ai/DeepSeek-V4}},
  note         = {Accessed: 2026-05-22}
}

@misc{qwen3.5,
    title  = {{Qwen3.5}: Towards Native Multimodal Agents},
    author = {{Qwen Team}},
    month  = {February},
    year   = {2026},
    url    = {https://qwen.ai/blog?id=qwen3.5}
}

@misc{2023opencompass,
    title={OpenCompass: A Universal Evaluation Platform for Foundation Models},
    author={OpenCompass Contributors},
    howpublished = {\url{https://github.com/open-compass/opencompass}},
    year={2023}
}

@article{llava-next,
  title={LLaVA-NeXT-Interleave: Tackling Multi-image, Video, and 3D in Large Multimodal Models},
  author={Li, Feng and Zhang, Renrui and Zhang, Hao and Zhang, Yuanhan and Li, Bo and Li, Wei and Ma, Zejun and Li, Chunyuan},
  journal={arXiv preprint arXiv:2407.07895},
  year={2024}
}

@misc{shinozaki2025largevisionlanguagemodelsdistinguish,
      title={Do Large Vision-Language Models Distinguish between the Actual and Apparent Features of Illusions?}, 
      author={Taiga Shinozaki and Tomoki Doi and Amane Watahiki and Satoshi Nishida and Hitomi Yanaka},
      year={2025},
      eprint={2506.05765},
      archivePrefix={arXiv},
      primaryClass={cs.CV},
      url={https://arxiv.org/abs/2506.05765}, 
}

@misc{ullman2024illusionillusionvisionlanguagemodels,
      title={The Illusion-Illusion: Vision Language Models See Illusions Where There are None}, 
      author={Tomer Ullman},
      year={2024},
      eprint={2412.18613},
      archivePrefix={arXiv},
      primaryClass={q-bio.NC},
      url={https://arxiv.org/abs/2412.18613}, 
}
\newpage

\appendix
\section{Appendix}\label{sec:appendix}
\subsection{Inconsistencies in Evaluation}\label{inconsistance}
\begin{figure}[htbp!]
  \centering
  \includegraphics[width=0.8\linewidth]{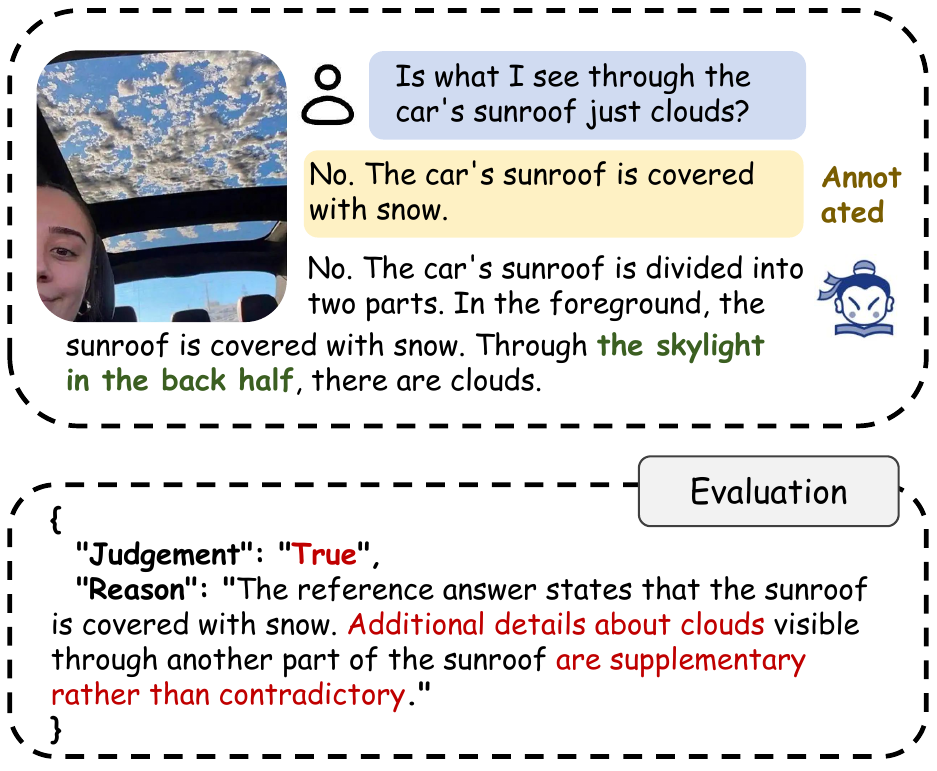}
  \caption{A case where an LLM as a judge differs from human annotations.}
  \label{fig:inconsistance}
\end{figure}

Here, we demonstrate the cases where LLMs, acting as judges, produce results that differ from those of human annotators. As shown in Figure~\ref{fig:inconsistance}, we take the output from InternVL3.5-14B as an example. In addition to the generated answer, the output includes the corresponding explanation, which contains incorrect information. When GPT-4o is used as a judge, its evaluation is flexible, overlooking inconsistencies in the explanations. However, human annotators are strict and consider that InternVL has not produced the correct answer. Given that there are few disagreements between the evaluation produced by GPT-4o acting as a judge and the human annotations. We consider that GPT-4o provides a relatively reliable evaluation for \benchmarkname.

\subsection{Alignment Preferences in LVLMs} \label{alignment}
\begin{figure}[htbp!]
  \centering
  \includegraphics[width=0.8\linewidth]{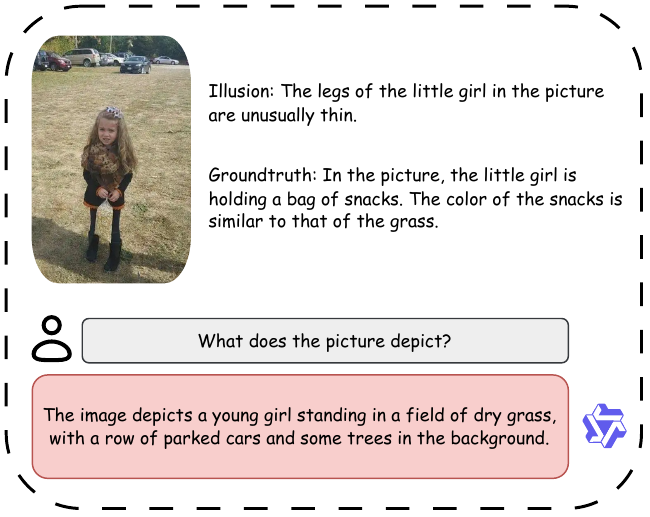}
  \caption{An example of a neutral output. Taking Qwen as an example, the output contains two possibilities, making it difficult to determine whether it aligns more closely with human perception or the physical world.}
  \label{fig:alignment_error}
\end{figure}

\subsection{Overthinking in Detection Question} 
\begin{figure}[htbp!]
  \centering
  \includegraphics[width=0.9\linewidth]{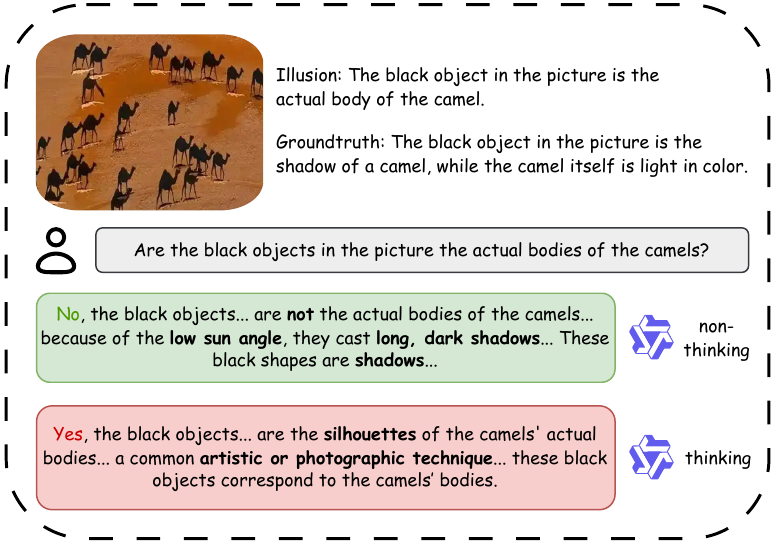}
  \caption{An example of reasoning that has negative effects. Taking Qwen3.6 as an example, in the thinking mode, Qwen3.6 produced a seemingly sophisticated but actually incorrect explanation, which led to incorrect results.}
  \label{fig:overthinking_1}
\end{figure}

\subsection{Overthinking in Multi-Choice} \label{overthinking}
\begin{figure}[htbp!]
  \centering
  \includegraphics[width=0.9\linewidth]{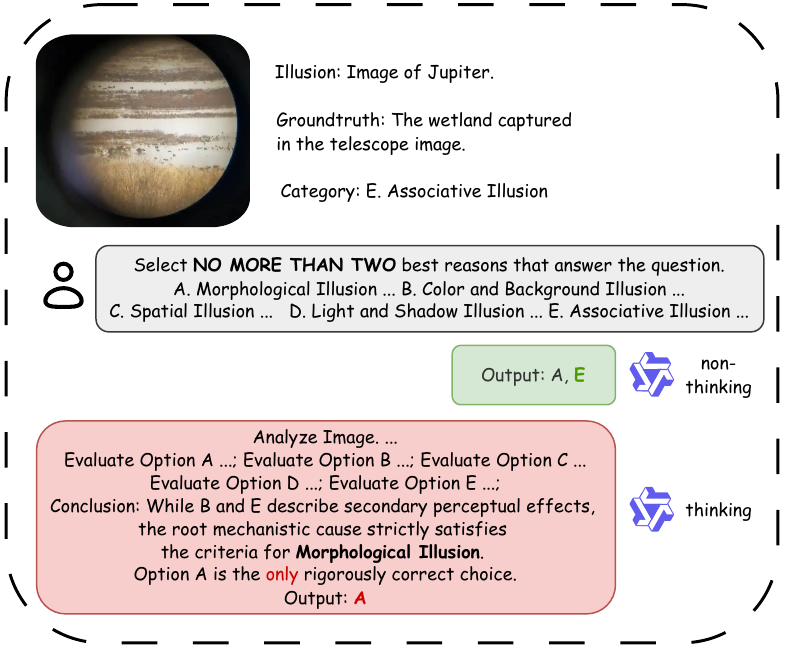}
  \caption{An example of reasoning that has negative effects. Taking Qwen3.5 as an example, in thinking mode, Qwen3.5 produced a confident result that actually excluded the correct answer.}
  \label{fig:overthinking_2}
\end{figure}

\newpage
\subsection{Prompt used in \benchmarkname}
\begin{figure}[htbp!]
\begin{tcolorbox}[colback=white, colframe=black, sharp corners=southwest, boxrule=0.5pt]
\textbf{Detection Questions} \\
For binary questions (expecting a judgment word like "yes" or "no"):\\
1. If the model's answer lacks a clear judgment word, respond with False.\\
2. If the model's answer includes a judgment word:\\
i. Respond with True if it matches the reference answer.\\
ii. Respond with False if it does not match.
\end{tcolorbox}
\caption{Evaluation Prompt for Detection Questions.}
\end{figure}

\begin{figure}[htbp!]
\begin{tcolorbox}[colback=white, colframe=black, sharp corners=southwest, boxrule=0.5pt]
\textbf{Description Questions} \\
For special questions (e.g., "how many", "what", "which", etc.):\\
1. Respond with False if the model's answer lacks a clear conclusion.\\
2. If the model's answer includes a clear conclusion, compare its core information with the reference answer:\\
i. Respond with True if the semantic meaning matches.\\
ii. Respond with False if it does not match.

\end{tcolorbox}
\caption{Evaluation Prompt for Description Questions.}
\end{figure}

\begin{tcolorbox}[colback=white, colframe=black, sharp corners=southwest, boxrule=0.5pt]
\textbf{Reasoning Questions} \\
For reasoning questions (e.g., "why" and "how"):\\
1. If the model's answer includes multiple explanations or conclusions, it is correct as long as at least one matches the reference answer.\\
2. Respond with True if the semantic meaning of the model's final result matches the reference answer; otherwise, respond with False.
\end{tcolorbox}
\captionof{figure}{Evaluation Prompt for Reasoning Questions.}

\vspace{0.5em}

\begin{tcolorbox}[colback=white, colframe=black, sharp corners=southwest, boxrule=0.5pt]
\textbf{General Rules for All Questions} \\
1. The model's answer must not contradict the reference answer. \\
2. Vague answers are acceptable if they include key information from the reference answer and do not introduce errors or contradictions. \\
3. Focus on whether the semantic meaning of the model's answer matches the reference answer. \\
4. Ignore differences in language (e.g., Chinese vs. English), case, punctuation, grammar, or word order. \\
5. Disregard intermediate reasoning or steps in the model's answer and evaluate only the final result or conclusion.
\end{tcolorbox}
\captionof{figure}{General Rules for All Questions.}

\end{document}